\documentclass[letterpaper, 10 pt, conference]{ieeeconf}  

\IEEEoverridecommandlockouts                              

\usepackage{graphics} 
\usepackage{epsfig} 
\usepackage{times} 
\usepackage{amsmath} 
\usepackage{amssymb}  
\usepackage{amsfonts}
\usepackage{amsmath}
\usepackage{cases}
\usepackage{algorithm}
\usepackage[noend]{algpseudocode}
\let\labelindent\relax
\usepackage{enumitem}
\usepackage{algorithm}
\usepackage{algpseudocode}
\usepackage{hhline}
\usepackage{graphicx}
\usepackage[OT1]{fontenc}
\usepackage[caption=false,font=footnotesize]{subfig}
\usepackage{booktabs}
\usepackage{multirow}
\usepackage{makecell}
\usepackage{array}

\title{\LARGE \bf
Designing Underactuated Grippers with Dynamically Variable Geometry Using Potential Energy Map Based Analysis
}

\author{C. L. Yako$^{1}$, Shenli Yuan$^{1}$, and J. Kenneth Salisbury$^{1}$
\thanks{$^{1}$Stanford Artificial Intelligence Lab
(SAIL), Stanford University}%
\thanks{\tt{clyako@stanford.edu, shenliy@stanford.edu, kenneth.salisbury@gmail.com}}
}

\begin{document}

\maketitle
\thispagestyle{empty}
\pagestyle{empty}

\begin{abstract}

This paper introduces an extension to the energy map method for in-hand manipulation.
Energy maps are used to predict how a part will evolve in the grasp given a specific actuation input to the gripper.
Previous approaches assumed frictionless contacts, but we show analytically that friction can be included in the energy maps when using two-link underactuated fingers by understanding the evolution of the part-finger contact.
These friction-based energy maps were used to evaluate the importance of various tendon-pulley gripper parameters across nearly 6 million simulated grasping scenarios. 
Specifically, a variable palm width is needed to manipulate parts of varying scales, and a variable transmission ratio, or the ratio of the distal to the proximal pulley radii, is needed to draw parts into a cage or to maintain a tip prehension grasp.

\end{abstract}

\section{INTRODUCTION}
\label{section: introduction}
One of the most daunting challenges still facing the robotics community is the design and control of robot grippers. 
A practical robot must be able to reliably grasp and manipulate in-hand a wide range of parts and yet be simple to use. 
Currently, most research focuses on the algorithmic side of grasping and in-hand manipulation systems, however, these are not one-dimensional problems; a gripper will only be as capable and dexterous as its hardware allows~\cite{chen2020hardware, ha2020fit2form}. 
We believe there is a middle ground that considers both the gripper design and corresponding control algorithm. 
To that end, this work involves the use of underactuated grippers and a control strategy referred to as energy maps. 

The underactuation affords the gripper a degree of compliance---it can automatically conform to the part shape without requiring complex control schemes~\cite{birglen2007underactuated,hirose1978development}. 
While this is beneficial for grasping, fine manipulation tasks are more difficult to achieve with these grippers.
Energy maps are a promising tool to overcome this challenge.
Bircher~\cite{bircher2019design, bircher2021complex} used these energy maps to manipulate a grasped part by driving it from high to low potential energy configurations.
The distribution of energy changes given different initial poses of the part (relative to the gripper) and / or actuation commands, hence, the part can intentionally be driven to various poses in the configuration space.
However, because of the frictionless assumption in previous approaches the manipulation was constrained to occur while parts were caged, limiting the part-gripper configuration space.

In this work, we are able to extend the use of energy maps to include friction.
The addition of friction relaxes the caging assumption, and allows tip-prehension grasps to be considered stable configurations.
The analysis presented here is valid for two-link tendon-pulley underactuated fingers, and can be extended to other two-link underactuated fingers. 
We first describe the analytical method in Section~\ref{section: methods}.
Then, in Section~\ref{section: experiments} we use this method to simulate nearly 6 million grasping tests across a range of part radii and contact conditions (i.e., coefficient of friction at contact). 
Section~\ref{section: discussion} discusses the results of these experiments: the design of grippers best suited to draw in various parts to a cage, the design of caging focused versus tip-prehension focused grippers, and the manipulation capabilities of a gripper that modulates its geometry versus one that modulates its tendon tension.
The third experiment was designed to investigate if the geometry-dependent reconfiguration capability of underactuated fingers could be used to manipulate parts by intentionally changing the finger geometry.

\section{METHODS}
\label{section: methods}
\begin{table}[tb]
\centering
    \caption{Nomenclature}
    \label{tab: gripper nomenclature}
    \begin{tabular}{c m{6.5cm}}
        \toprule
        \textbf{Symbol(s)} & \textbf{Meaning}\\
        \midrule 
        $N_o$ & origin of the world frame $N$, fixed to the center of the gripper palm\\
        $\hat{n}_x$ / $\hat{n}_y$ & world frame unit vectors \\
        $A_o$ & origin of frame $A$, fixed to the proximal joint\\
        $B_o$ & origin of frame $B$, fixed to the distal joint\\
        $\hat{a}_x$ & unit vector pointing from $A_o$ to $B_o$\\
        $\hat{a}_y$ / $\hat{b}_y$ & unit vector defining the contact normal direction for forces from the gripper on the part\\
        $\hat{b}_x$ & unit vector pointing from $B_o$ to the tip of the distal link\\
        $l_i$ & length of link $i$\\
        $k_i$ & distance from joint $i$ to the contact location on link $i$\\
        $w$ & width of the gripper palm\\
        $r_i$ & pulley radius at joint $i$\\
        $\theta_1$ & angle from $-\hat{n}_x$ to $\hat{a}_x$ with $+\hat{n}_z$ sense (left) or from $\hat{n}_x$ to $\hat{a}_x$ with $-\hat{n}_z$ sense (right)\\
        $\theta_2$ & angle from $\hat{a}_x$ to $\hat{b}_x$ with $+\hat{n}_z$ sense (left) or with $-\hat{n}_z$ sense (right)\\
        $F_i$ & magnitude of normal contact force on link $i$\\
        $\mu_s$ & static coefficient of friction\\
        $\mu_i$ & multiplier between $\pm\mu_s$ used to determine tangential contact force on link $i$\\
        $F_{a}$ & tendon tension\\
        $K_i$ & spring constant for joint $i$\\
        $r$ & radius of the part\\
        $\Vec{p}$ & position vector from $N_o$ to the center of the part\\
        \bottomrule
    \end{tabular}
\end{table}
\begin{figure}[ht!]
  \centering
  \includegraphics[width=0.48\textwidth]{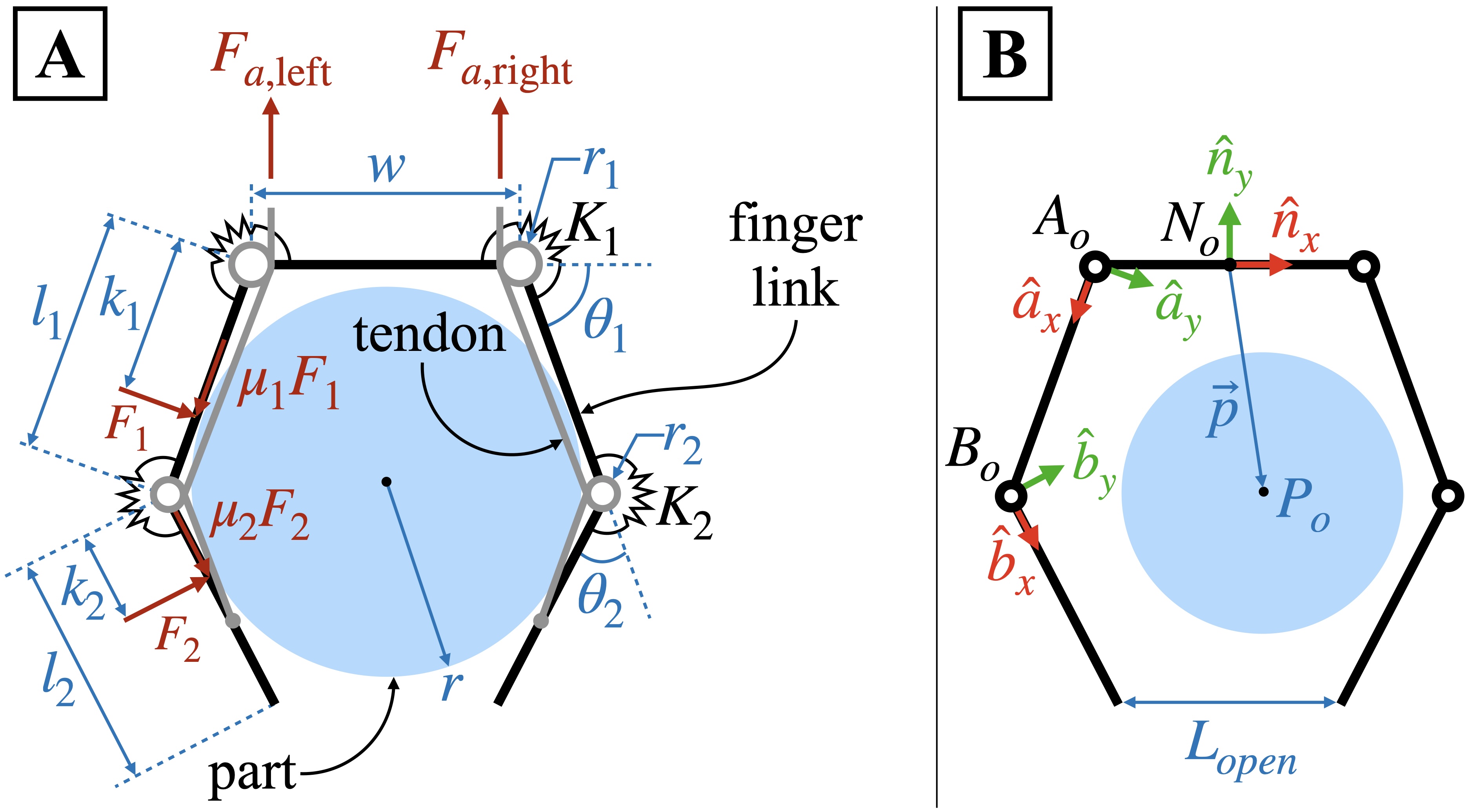}
    \caption[Description of the gripper-part system used in creating the friction-based energy maps.]{(A) Geometric parameters as well as contact forces on the part and finger actuation forces. 
    This chapter assumes a tendon-pulley underactuated gripper, with the tendon shown in gray and the zero-thickness gripper links shown in black. 
    Springs at the joints, also shown in black, enforce a proximal-to-distal closing sequence of the gripper, but create negligible joint torques relative to the applied torques.
    (B) Frames for each link as well as the world frame, described by $\hat{n}_x$, $\hat{n}_y$ and origin $N_o$.
    The center of the palm is fixed to $N_o$. 
    The position vector $\Vec{p}$ describes the position of the center of the part, $P_o$, from $N_o$. 
    Frame $A$ is fixed to the proximal link with its origin, $A_o$ coincident with the proximal joint. 
    The same is true of frame $B$ but for the distal link and joint.}
    \label{figure: gripper parameters}
\end{figure}
In this section, we first discuss the assumed gripper model. 
We then describe how the configuration of the fingers can be determined for any part-gripper position, and how this is related to energy. 
Finally, we discuss the construction of a given energy map. 
The method presented here for determining the potential energy of a given part-gripper configuration is specific for two-phalanx tendon-driven fingers, and the approach can be adapted based on the mechanism~\cite{birglen2007underactuated}.
The analysis of the potential energy maps presented can be used for any planar gripper.
\vspace{-1mm}

\subsection{Kinematic Model}
The geometric parameters, forces, and system frames are shown in Fig.~\ref{figure: gripper parameters}, and variable descriptions can be found in Table~\ref{tab: gripper nomenclature}.
The part being grasped is assumed to have a radius of $r$.
The left and right fingers are symmetric in design: link lengths $l_1$ and $l_2$, pulley radii $r_1$ and $r_2$, and joint springs $K_1$ and $K_2$. 
The tendon tensions, applied through the gray cable, are $F_{a,\text{left}}$ and $F_{a,\text{right}}$ for the left and right fingers. 
We will refer to the tendon tension not specific to a certain finger as $F_a$. 
Given a nonzero $F_a$, the torques at each joint will be equal to $T_1 = F_a r_1$ and $T_2 = F_a r_2$. 
In response to these torques the finger will move until the contact forces with a given part result in static equilibrium for the finger. 
We assume that the spring torques, given by $K_1 \theta_1$ and $K_2 \theta_2$, keep each phalanx at its lower joint limit during free space motion, and are negligible relative to both the joint torques and the moments applied by the grasped part. 
However, the springs are necessary to resolve the free space motion indeterminacy of the finger. 
Specifically, we assume they compel the proximal $\rightarrow$ distal movement behavior described by Hirose~\cite{hirose1978development}, i.e., the distal phalanx will not move until the proximal phalanx reaches a joint limit or makes contact with the part.
The center of the palm is fixed to the origin of the world frame $N$ at $N_o$.
Frames $A$ and $B$ are fixed to the proximal and distal links with their origins coincident with the proximal and distal joints.

\subsection{Potential Energy Maps with Friction}
Potential energy maps aim to describe the gross motion of a part given a specific actuator command and an initial part-gripper position, defined by $\Vec{p}$ in Fig.~\ref{figure: gripper parameters}. 
These maps are constructed by first fixing the part in place, then determining the configuration of the fingers that result in static equilibrium of the gripper as well as the energy of the resulting stable system. 
Once the energy has been determined for the entire configuration space of the part and gripper, the fixed part can be released, and its motion can be inferred from the gradient of the energy map~\cite{bircher2021complex}.

In a frictionless system, the equilibrium configuration can be found by minimizing the energy of each finger. 
From Kragten~\cite{kragten2011underactuated}, the actuators are the sole energy source and thus the potential energy can be measured by the work done by the joint torques, $W_{actuators}$, and spring torques, $W_{springs}$. 
For a finger with $n$ joints, the joint variable vector $q = [\theta_{1},...,\theta_{n}]$, and springs at each joint with spring constant $K_i$ the equilibrium joint angles of the finger can be found by carrying out the minimization in~\eqref{eq: energy minimization no friction} subject to the appropriate contact constraints.
Note that $\theta_{i_o}$ is the rest position for each joint.

\begin{align}\label{eq: energy minimization no friction}
    \min_{q} V &= -W_{actuators} + W_{springs} \notag\\
    &= -\sum_{i=1}^{n} T_{i} (\theta_{i} - \theta_{i_o}) - \frac{1}{2}K_i (\theta_{i}^2 - \theta_{i_o}^2)
\end{align}

By solving~\eqref{eq: energy minimization no friction} for each part-gripper configuration, we can predict the gross motion of the part by following energy gradients from high potential energy to low potential energy configurations. 
This was experimentally shown by Bircher et al.~\cite{bircher2019design, bircher2021complex}.

\subsubsection{Stable part-gripper positions with friction}
A benefit of continuing to use energy to predict gross part motion is that, similar to conservative systems, non-conservative systems tend to move from high energy to low energy part-gripper positions. 
\begin{figure}[b!]
    \centering
    \includegraphics[width=0.48\textwidth]{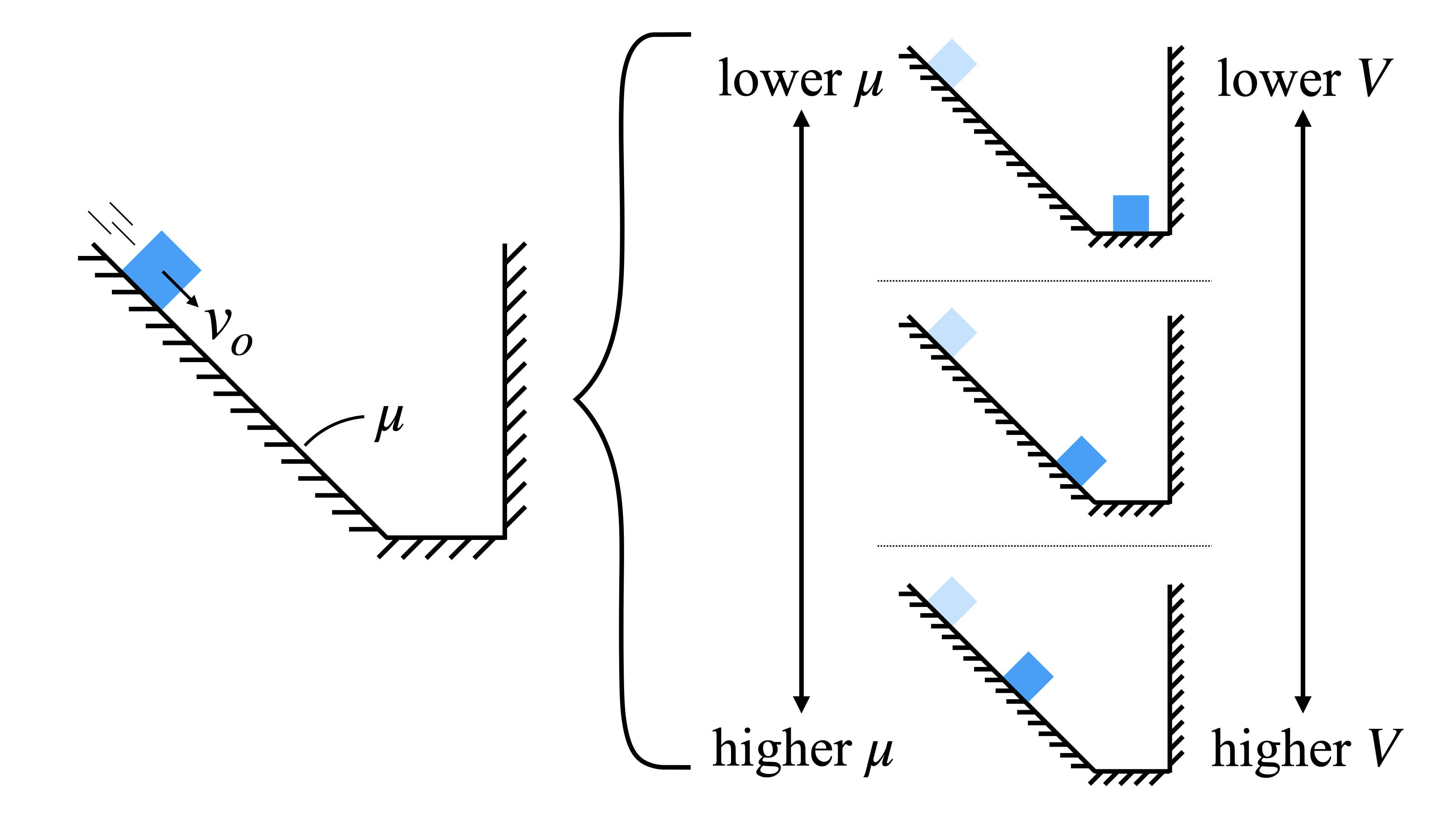}
    \caption[Example demonstrating how systems with friction still tend toward lower energy states.]{The blue box has initial velocity, $v_o$.
    Friction, $\mu$, prevents the box from reaching the minimum possible potential energy state, $V$, though the system tends towards lower energy states.}
    \label{figure: energy example}
\end{figure}
However, the inclusion of friction indicates the system may not reach the absolute lowest energy state, but may instead settle at a different, higher energy state where static equilibrium is also achieved. 
This is analogous to a box with initial velocity $v_o$ sliding down a ramp with friction, where friction prevents the box from reaching the lowest possible energy state, as shown in Fig.~\ref{figure: energy example}.
The same is true of our part-gripper system.

In this work, we consider rotationally symmetric parts and therefore static equilibrium states are achieved when the convex hull in force space defined by the friction cone edges at each contact point contain the external force, $F_{ext}$. 
This is shown in~\eqref{eq: contact force hull}, where $G^* \in \mathcal{R}^{2 \times 4N_c}$ is the force-only grasp matrix for a given finger configuration, $N_c$ is the number of contact points, and $c^T = [c_1, ..., c_{2N_c}]$ is the vector of friction cone edges. 
For the $i$th contact, where $i \leq 2N_c$, we have $c_i = [F_{t_i}, F_{n_i}]$ consisting of a tangential force, $F_{t_i}$, and a normal force, $F_{n_i}$
We assume a hard point contact with Coulomb friction, therefore $|F_{t_i}| \leq \mu_s F_{n_i}$, where $\mu_s$ is the coefficient of static friction.
\begin{equation}\label{eq: contact force hull}
    G^*c = -F_{ext}
\end{equation}
In our case we assume that external forces on the part are negligible relative to the applied forces, i.e., $F_{ext}$ is the zero vector. 
Before we can use~\eqref{eq: contact force hull} to see if we are at a stable part-gripper position, we need to determine the configuration of fingers that result in static equilibrium of the gripper. 

\subsubsection{Finger configuration determination}\label{sec: fingerConfigruationDetermination}
The static equilibrium configuration of each finger is needed in order to calculate the potential energy of the system (see Section~\ref{sec: energyEquation}). 
We cannot simply use~\eqref{eq: energy minimization no friction} to find the configuration (or system potential energy) because the presence of friction implies multiple finger configurations satisfying static equilibrium. 
Instead, we need to understand the evolution of the grasp by considering both the vector loop equations and the sign of the proximal contact force~\cite{birglen2007underactuated}. 
First, a least-squares minimization is performed on the vector loop equations for the proximal link~\eqref{eq: proximal vector loop} and the distal link~\eqref{eq: distal vector loop}.
If no solution is found for~\eqref{eq: proximal vector loop} and~\eqref{eq: distal vector loop}, then the part is not reachable. 

\begin{align}
    \Vec{p} &= 
    \begin{bmatrix} \pm w / 2 \\ 0\\ \end{bmatrix}  +~^N_AR(\theta_1)
    \begin{bmatrix} k_1 \\ r\\ \end{bmatrix} 
    \label{eq: proximal vector loop}\\
    \Vec{p} &= 
    \begin{bmatrix} \pm w / 2 \\ 0\\ \end{bmatrix}  +~^N_AR(\theta_1)
    \begin{bmatrix} l_1 \\ 0\\ \end{bmatrix} +~^N_BR(\theta_1, \theta_2) \begin{bmatrix} k_2 \\ r\\ \end{bmatrix}
    \label{eq: distal vector loop}
\end{align}

Next, the sign of the proximal contact force, $F_1$ is investigated using~\eqref{eq: normal contact proximal link} and~\eqref{eq: normal contact distal link}. 
It is necessary to evaluate~\eqref{eq: normal contact proximal link} for $|\mu| \leq \mu_s$ as shown in~\eqref{eq: case 1} -~\eqref{eq: case 3}, since the magnitude of the tangential force on the distal link must be less than or equal to $\mu_s F_2$. 
Given a solution to both~\eqref{eq: proximal vector loop} and~\eqref{eq: distal vector loop} static equilibrium is achieved if $F_1$ is strictly positive (case~\eqref{eq: case 1}), or if we can achieve $F_1 = 0$ with $|\mu| < \mu_s$ since we have sticking (case~\eqref{eq: case 3}). 
If $F_1$ is strictly negative (case~\eqref{eq: case 2}), then the contact point will slide toward the tip of the distal link until static equilibrium is reached or the part is ejected. 

\begin{align}
    F_{1} &= \frac{F_a r_1 - (l_1\cos(\theta_2) + k_2)F_{2} + l_1 \sin(\theta_2) \mu F_{2}}{k_1} \label{eq: normal contact proximal link}\\
    F_{2} &= \frac{F_a r_2}{k_2} \label{eq: normal contact distal link}
\end{align}

A solution to only~\eqref{eq: proximal vector loop} implies the proximal link is contacting the part and the distal link has reached its joint limit, thus the sign of $F_1$ does not need to be investigated.
If only~\eqref{eq: distal vector loop} is satisfied, then only the distal link is able to contact the part. 
Then cases~\eqref{eq: case 1} and~\eqref{eq: case 3} imply static equilibrium has been achieved, while case~\eqref{eq: case 2} implies reconfiguration.

\begin{subnumcases}{\parbox{1.75cm}{\centering\textit{proximal\\link stick-\\slip cases}}}
    F_1 > 0, & $\forall\mu \text{, where } |\mu| \leq \mu_{s}$ \label{eq: case 1}\\
    F_1 < 0, & $\forall\mu \text{, where } |\mu| \leq \mu_{s}$ \label{eq: case 2}\\
    \exists! \mu \text{ s.t. } F_1 = 0, & $\quad \:\: \text{ where } |\mu| < \mu_{s}$ \label{eq: case 3}
\end{subnumcases}

\subsubsection{System potential energy}\label{sec: energyEquation}
We consider the system as the gripper and part.
As described above, the system potential energy is the work done by the joint torques, $W_{actuators}$, and the spring torques, $W_{springs}$~\cite{kragten2011underactuated}.

\begin{equation}
    V_{system} = -W_{actuators} + W_{springs} \label{eq: energy from actuator and springs}
\end{equation}

Friction is internal to this system, relating to the flow of energy as heat and changes of internal energy of both the gripper and part, at least on the timescale of the gripping action. 
Thus, it is not explicitly included in~\eqref{eq: energy from actuator and springs}. 
However, while it may seem that friction has no effect on $V_{system}$, it is implicitly included in~\eqref{eq: energy from actuator and springs} as it affects the equilibrium joint angles, which in turn affect both $W_{actuators}$ and $W_{springs}$. 
As expected, given two identical scenarios (same gripper and part position) that differ only in the value of $\mu_s$, the scenario with the larger $\mu_s$ will have a higher value $V_{system}$.

Just as in~\eqref{eq: normal contact proximal link} and~\eqref{eq: normal contact distal link} we assume that the springs are negligible and so energy is calculated by only considering the work done on the joints for each finger (left and right) as was done by Kragten~\cite{kragten2011underactuated}:
    
\begin{equation}
    V_{system} = \sum_{l,r}\sum_{i=1}^{2} \left(-T_{i} (\theta_{i} - \theta_{i_o})\right) \label{eq: final energy equation}
\end{equation}

\begin{figure}[tb]
    \centering
    \includegraphics[width=0.50\textwidth]{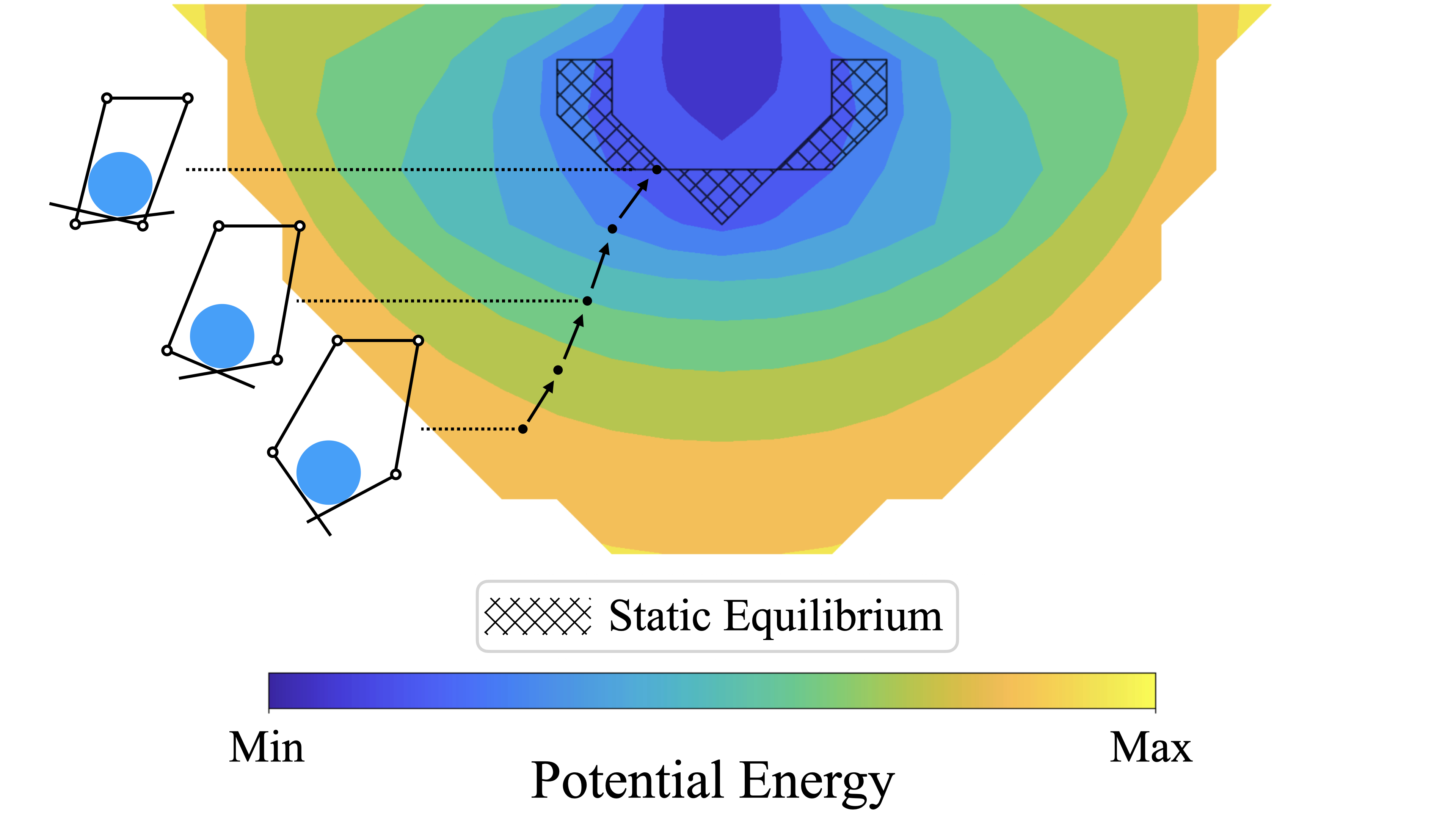}
    \caption[Potential energy map with an example part trajectory.]{Potential energy map. Hatched regions indicate static equilibrium, i.e.,~\eqref{eq: contact force hull} is satisfied. 
    Parts will tend to move from high potential energy to low potential energy configurations, as indicated by the black arrows. 
    Friction may prevent the lowest possible energy state from being reached. 
    Note that the gripper and part are not to scale with the energy map, but are there to illustrate the evolution of the grasp.}
    \label{figure: energy map}
\end{figure}

\subsubsection{Potential energy map construction}

To construct a potential energy map, we first fix the gripper, and then place the part at various grid points $\mathcal{P}$. 
Energy map $E_j$ is constructed by plotting the energy found from~\eqref{eq: final energy equation} at each point $\Vec{p} \in \mathcal{P}_j$, where $\mathcal{P}_j \subseteq \mathcal{P}$ and is the subset of points where the part could be reached.
A representative energy map can be seen in Fig.~\ref{figure: energy map} -- the hatched areas in Fig.~\ref{figure: energy map} are locations where the force equilibrium component of~\eqref{eq: contact force hull} is satisfied and we therefore assume the part will not translate. 

Energy maps can be compared depending on the performance requirements of the gripper. 
One such requirement could be a gripper's ability to evolve a part into a stable cage or stable tip prehension point, or both. 
To analyze this, we can take each graspable point and follow the gradient of the energy into one of the following types of possible end states:

\begin{subnumcases}{\parbox{1.5cm}{\centering\textit{possible\\end states}}}
    \textbf{ejection: }~\eqref{eq: proximal vector loop} -~\eqref{eq: normal contact distal link} \text{ not satisfied}\nonumber\\
    \textbf{stable: }~\eqref{eq: contact force hull} -~\eqref{eq: normal contact distal link} \text{ are satisfied}\nonumber\\
    \textbf{caged: } \text{ stable, } L_{open} < 2r, \text{ contained}\nonumber
\end{subnumcases}

If a part is said to be ``contained" it means that the center of the part is within the polygon formed by the gripper links and palm and $L_{open}$ (see Fig.~\ref{figure: gripper parameters}B) describes the largest possible part that could escape the grasp~\cite{bircher2019design}. 
Points where~\eqref{eq: proximal vector loop} -~\eqref{eq: normal contact distal link} are satisfied but~\eqref{eq: contact force hull} is not imply that the part is still evolving towards its final grasp, since we do not have force closure. 
In these cases, we assume the part is evolving along the path of least resistance (i.e., following energy gradients). 
Stable points consist of both caged and tip prehension grasps; we did not consider form closure grasps in our characterizations.

\subsection{Limitations}
The method discussed here has several limitations. 
Namely, it only works for 2-DoF \textit{underactuated} fingers, as the closing sequence of the finger is leveraged to resolve the indeterminacy from frictional contacts.
For an $n$-link finger with the same proximal to distal closing sequence, grasp stability still remains an open problem~\cite{birglen2007underactuated}.
There are also situations where the palm may apply forces on the part.
Assuming a hard point contact with friction, these forces are unknown. 
If there is no friction, then the normal force can be determined if there is distal-only contact on both fingers.
Note that checking if~\eqref{eq: contact force hull} is satisfied does not require values of these forces to be known.
Additionally, we ignored part rotation since we narrowed our analysis to rotationally symmetric parts.
However, this analysis can be extended by considering the entire planar wrench space.
Finally, we made the assumption that parts not in equilibrium will evolve along the energy gradient, though friction can lead to infinitely many possible paths.

\section{NUMERICAL EXPERIMENTS}
\label{section: experiments}
The goals of our experiments were as follows: 

\begin{itemize}
    \item Determine the best gripper designs defined by their ability to cage a specific part.
    \item Provide design insights with respect to gripper parameters for a variety of part sizes and friction coefficients. 
    \item Investigate the possible trade-off between caging and tip prehension grasps.
    \item Compare a gripper with dynamically variable geometry ($w$, $l_1$, $l_2$, $r_1$, and $r_2$ can all actively change) to a set of grippers with only a variable palm, with respect to a custom in-hand manipulation metric.
\end{itemize}

\begin{table}[t!]
\centering
\caption{\label{tab: parameterSearch} Parameters Used in the Design Space Search}
\small
\setlength{\tabcolsep}{4pt}
\begin{tabular}{lccccccc}
    \toprule
    \multirow[c]{4}{*}{\makecell[c]{\textbf{Design Search} \\ \textbf{Parameter}}} & 
    \multicolumn{5}{c}{\multirow[c]{2}{*}{\textbf{Gripper Geometry}}} & 
    \multirow[c]{2}{*}{\makecell[c]{\textbf{Part} \\ \textbf{Size}}} & 
    \multirow[c]{2}{*}{\makecell[c]{\textbf{Contact} \\ \textbf{Condition}}} \\[2.5ex] 
    \cmidrule(lr){2-8} 
    & $l_1$ & $l_2$ & $r_1$ & $r_2$ & $w$ & $r$ & $\mu_s$\\ 
    \midrule %
    \textit{Lower Bound} & 0.4 & 0.4 & 0.04 & 0.02 & 0.0 & 0.4 & 0.1\\
    \textit{Upper Bound} & 2.0 & 1.6 & 0.20 & 0.18 & 2.0 & 1.6 & 1.0\\
    \textit{Step Size}   & 0.2 & 0.4 & 0.04 & 0.04 & 0.4 & 0.4 & 0.3\\
    \bottomrule
\end{tabular}
\end{table}

\begin{figure}[b!]
    \centering
    \includegraphics[width=0.50\textwidth]{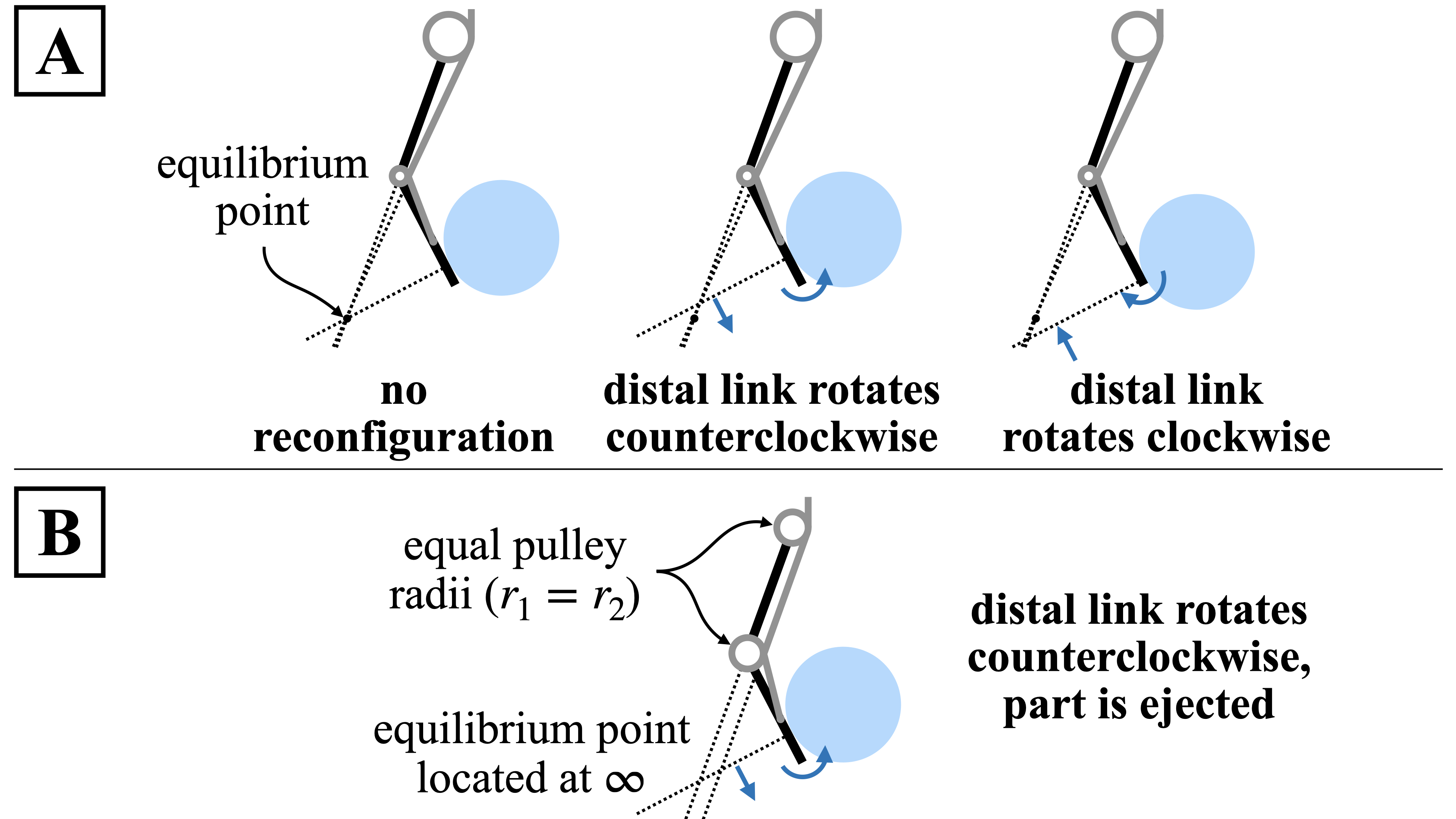}
    \caption[Geometric reasoning for reconfiguration grasps in distal-only contact scenarios.]{Geometric reasoning for reconfiguration grasps in distal-only contact scenarios.
    For simplicity's sake, we assume no friction, though it can be included by replacing the dotted line corresponding to the contact normal with two rays denoting the friction cone edges.
    (A) If the contact normal passes through the equilibrium point (left), which is the intersection of the tendon line and the proximal phalanx line, then the finger is in equilibrium. 
    Otherwise, the distal link will rotate according the moment the contact normal produces about the equilibrium point (center, right) until the contact normal intersects the equilibrium point.
    (B) If the pulley radii are equal, then the equilibrium point is located at infinity and rotation of the distal link will cause the part to be lost from the grasp.
    }
    \label{figure: equal pulley radii}
\end{figure}

To do this we analyzed the performance of 5,400 gripper designs on 16 parts, defined by the parameters shown in Table \ref{tab: parameterSearch}, resulting in a total of 86,400 energy maps.
Values for the parameters were unitless, and can be scaled up or down to physical dimensions as necessary. 
While the parameter ranges were empirically determined, specifically to capture designs that worked well and those that did not, our method can easily be expanded to any desirable ranges. 
No experiments were run with $r_1 = r_2$, because for reconfiguration grasps where two-phalanx contact shifts to single-phalanx contact, the contact point for static equilibrium would need to be located at infinity (Fig.~\ref{figure: equal pulley radii}).
The proximal joint limit was set to $0^\circ \leq \theta_1 \leq 180^\circ$ and the distal joint limit was set to $-5^\circ \leq \theta_2 \leq 90^\circ$. 
The 86,400 energy maps were computed in roughly 10 hours using a mix of 256 Intel Xeon E5-2640v4, Intel Xeon Gold 5118, and AMD 100 Epyc 7502 CPU cores on a high performance computing cluster.

\subsection{Caging Grasps}\label{sec: experimentsCaging}
We scored grippers for a given part by their ability to evolve the part along a trajectory into a caging grasp. 
Each finger was given the same equal actuation effort, i.e., $F_{a,\text{left}} = F_{a,\text{right}}$. 
Trajectories were generated like streamlines to indicate whether the part reached a static equilibrium configuration before an energy minimum.
Trajectories that ended in ejection or a tip prehension point were given a score of zero, while trajectories that ended in a cage were scored according to $L_{open}$.
The score for the $j$th energy map, $\Lambda_j$, was calculated using the caging metric~\eqref{eq: caging metric} from Bircher et al.~\cite{bircher2021complex}.
Given $n_{c,j}$ points in $\mathcal{P}_j$ that ended in a cage:
\begin{align}
    \label{eq: caging metric}
    S_i &= 1 - \frac{L_{open,i}}{2r}\\
    \label{eq: energy map caging score}
    \Lambda_j &=\sum_{i=1}^{n_{c,j}} S_i 
\end{align}

\subsection{In-Hand Manipulation Metric}\label{sec: metric}

We take in-hand manipulation to refer to intentionally changing the part pose relative to a frame fixed to the gripper palm.
Traditionally, parts are manipulated in-hand by commanding varied actuator efforts, or in our case varied tendon forces, which cause different resultant forces to be applied to the part as the finger configuration changes.
Different resultant forces can also be realized by directly varying the gripper geometry (link lengths, palm width, transmission ratios), even while maintaining a constant tendon tension.
Using these two methods of in-hand manipulation, which we define below as the ability of a gripper to apply an arbitrary force in any direction to a part in a given pose, we compared the in-hand manipulation capability of a gripper that varies its tendon tensions and has a variable palm against a gripper that \textit{only} varies its geometry (while maintaining a fixed tendon tension).
We leveraged the method of constructing the friction based potential energy maps, using the known grasping sequence to establish the final finger configuration, to determine the resultant forces on a part.

The manipulation metric used here is nearly identical to that described by Bircher et al.~\cite{bircher2021complex}, with a few subtle differences. 
We first fix a given part at a specific location.
Then, the set of contact forces for a specific set of actuation commands and for a specific gripper configuration (i.e., gripper geometry) are determined, scaled by the caging score~\eqref{eq: caging metric}.
If force equilibrium of the part~\eqref{eq: contact force hull} is satisfied, then these contact forces are discarded since no motion will be imparted to the part.
This is repeated across the set of all actuation commands and allowable gripper configurations.
For this specific part position, we construct the convex hull using the previously determined set of contact forces for all actuation commands and allowable gripper configurations, and then calculate the radius of the largest inscribed circle.
This is a measure of the largest force that can be imparted to the part in any direction.
We repeat this procedure across all part poses and sum the radii for each pose---this value is the manipulation metric for a given gripper and part.
Note that configurations where the part was making contact with the non-frictionless palm were discarded, since these palm-part contact forces are unknown.

We considered two scenarios:
\begin{itemize}
    \item {\textit{Scenario A}}
    \begin{itemize}
        \item {Gripper: Variable $w$, fixed $l_1$, $l_2$, $r_1$, and $r_2$}
        \item {Actuation: $F_{a,\text{left}} + F_{a,\text{right}} = \alpha$}
    \end{itemize}
    \item {\textit{Scenario B}}
    \begin{itemize}
        \item {Gripper: Variable $w$, $l_1$, $l_2$, $r_1$, and $r_2$}
        \item {Actuation: $F_{a,\text{left}} + F_{a,\text{right}} = \alpha,\text{with } F_{a,\text{left}} = F_{a,\text{right}}
        $}
    \end{itemize}
\end{itemize}

The grippers for \textit{Scenario A} were sampled from the best caging gripper designs determined from the experiments described in \ref{sec: experimentsCaging}. 
\textit{Scenario B} assumes a gripper that can realize all 5,400 designs in Table \ref{tab: parameterSearch}. 
We set $\alpha = 2.0$ and sample each finger force, $F_{a,\text{left}}$ and $F_{a,\text{right}}$, in steps of $0.2$ from $0.2$ to $1.8$. 
Note that in \textit{Scenario B} there is only one set of actuation forces ($F_{a,\text{left}} = F_{a,\text{right}} = 1.0$), namely a set of forces specifically geared for pulling the part toward the palm. 
We computed the described manipulation metric for each gripper in \textit{Scenario A} and compared the best design to that for \textit{Scenario B} for all sixteen parts described in Table~\ref{tab: parameterSearch}.

\section{DISCUSSION}
\label{section: discussion}
\begin{figure*}[tb]
    \centering
    \includegraphics[width=0.95\textwidth]{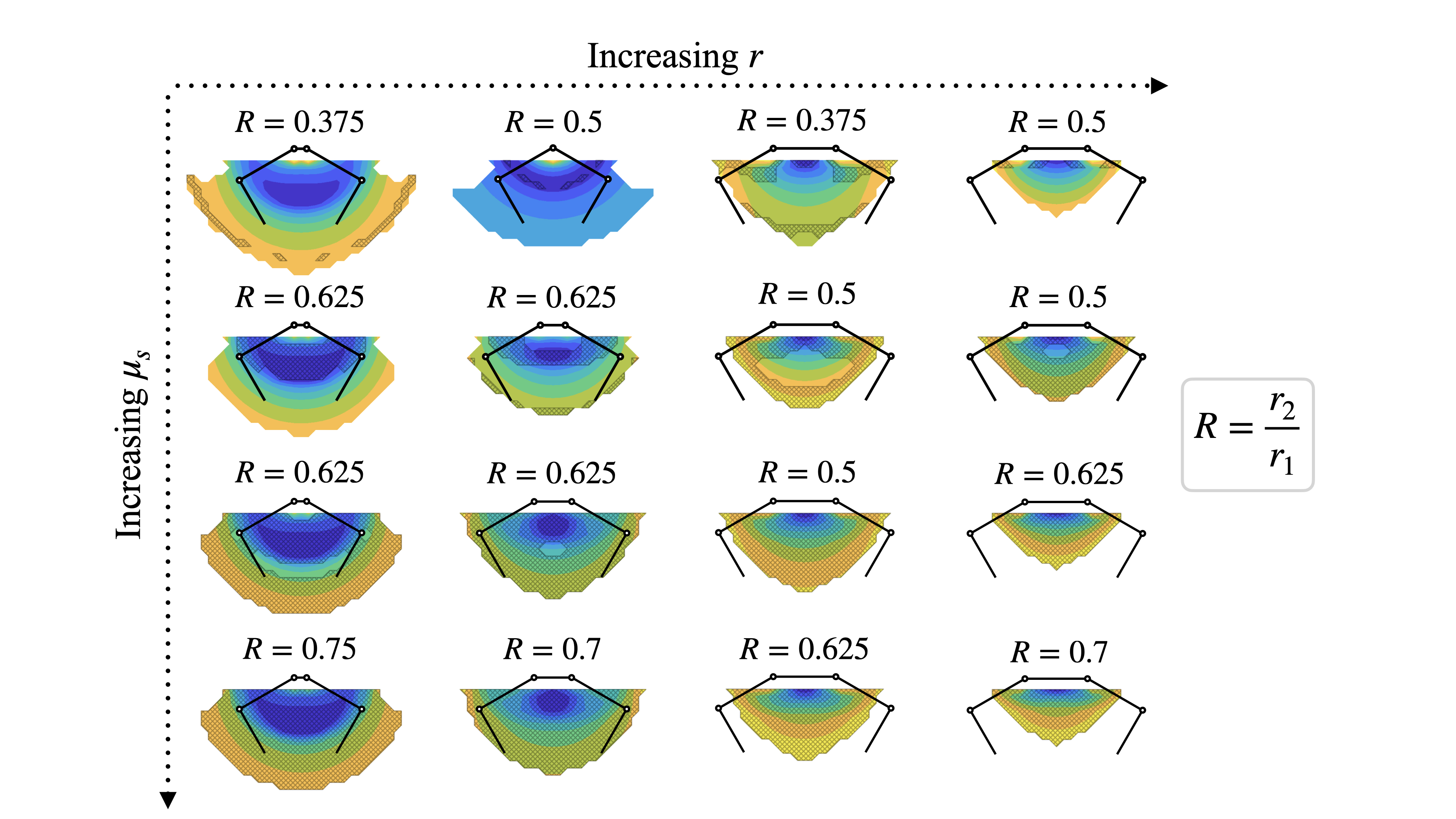}
    \caption[Best grippers at evolving a given part, defined by $r$ and $\mu_{s}$, into a caging grasp along with the corresponding energy maps.]{Best grippers at evolving the given part, defined by $r$ and $\mu_{s}$, into a caging grasp.
    Each gripper is shown to scale on top of its corresponding energy map for the given part. 
    Darker colors represent lower energy states for the system, and hence the part will tend to be pulled into these regions.}
    \label{figure: best caging designs}
\end{figure*}
\subsection{Caging Ability}
The best gripper design at establishing a caging grasp, along with its corresponding energy map for each of the sixteen parts tested, can be seen in Fig.~\ref{figure: best caging designs}. 
Note that the energy maps appear to be relatively smooth, i.e., a small change in a gripper design may lead to a small change in the energy map. 
Therefore, a sparser search could be performed in the gripper design space, and optimal designs could be found through interpolation in energy map space.
For the grippers below, each design maximizes its finger length, such that $l_1 = 2.0$ and $l_2 = 1.6$ (the upper bounds of our parameter search). 
Relative to a gripper with shorter links, longer fingers allow the gripper to reach the given part in more configurations, and to reach further around the part in order to better pull it towards the palm. 
Both of these will increase $\Lambda_j$ by increasing $n_{c,j}$ and decreasing $L_{open,i}$. 

There are two additional trends to note, which can be seen by looking along the columns (constant part radius $r$) and along the rows (constant $\mu_{s}$) in Fig.~\ref{figure: best caging designs}. 
In the first column of Fig.~\ref{figure: best caging designs}, the palm width is equal to the part radius ($w = 0.4$), and as the part size increases so does the palm width ($0.8 \rightarrow 1.2$ in three of the four cases in the third column, and $2.0$ in the right two columns).
If the palm is too small, the fingers may not be able to get around the part and pull it into a caging grasp (rightmost columns). 
On the other hand, both fingers may not be able to reach the part if the palm is too large (leftmost columns). 

For a given part radius $r$, the optimal transmission ratio ($R = r_2 / r_1$) increases as the coefficient of static friction increases. 
This relationship can be understood by noting the importance of $k_2$. 
As shown in Fig.~\ref{figure: contact sliding}, as the distal link rotates, $k_2$ increases and the contact point moves toward the fingertip. 
This re-positioning allows the finger to slide itself further underneath the part causing the contact normal to be directed more toward the palm, therefore making it easier to pull the part into a caging grasp. 
The effect of $R$ on $k_2$ can be seen by looking at the value of $k_2$ needed to achieve moment equilibrium about the proximal joint when there is distal only contact~\eqref{eq: distal equilibrium position}. 
In other words, if we set~\eqref{eq: normal contact proximal link} equal to zero and solve for $k_2$ in terms of $R$ we get the following:

\begin{figure}[t!]
    \centering
    \includegraphics[width=0.40\textwidth]{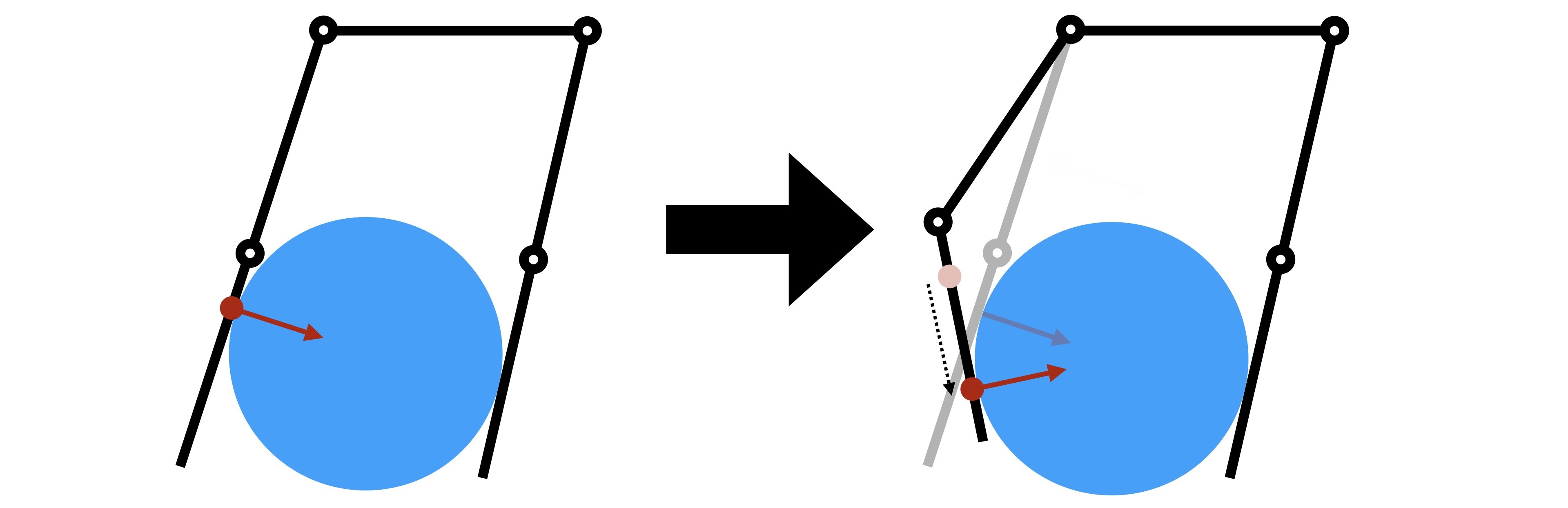}
    \caption[Contact point evolution on the distal link with a large transmission ratio, $R$.]{Contact point evolution on the distal link. 
    If the transmission ratio $R$ is sufficiently large, the contact point will slide toward the fingertip of the distal link, causing the contact normal to orient toward the palm.}
    \label{figure: contact sliding}
\end{figure}

\begin{align}
    k_2 &= \frac{R}{1 - R}l_1\left(\cos(\theta_2) - \mu_{s} \sin(\theta_2)\right) \label{eq: distal equilibrium position}
\end{align}

The transmission ratio $R$ and friction $\mu_{s}$ have countering effects: an increase in $\mu_s$ causes $k_2$ to decrease (undesirable for caging) and an increase in $R$ causes $k_2$ to increase (desirable for caging). 
There are limits, however, since we need $k_2 \leq l_2$ for a stable grasp, which constrains the transmission ratio $R$ from being excessively large. An alternative is to actively modulate the coefficient of friction of the contact. 

\begin{figure}[b!]
    \centering
    \includegraphics[width=0.48\textwidth]{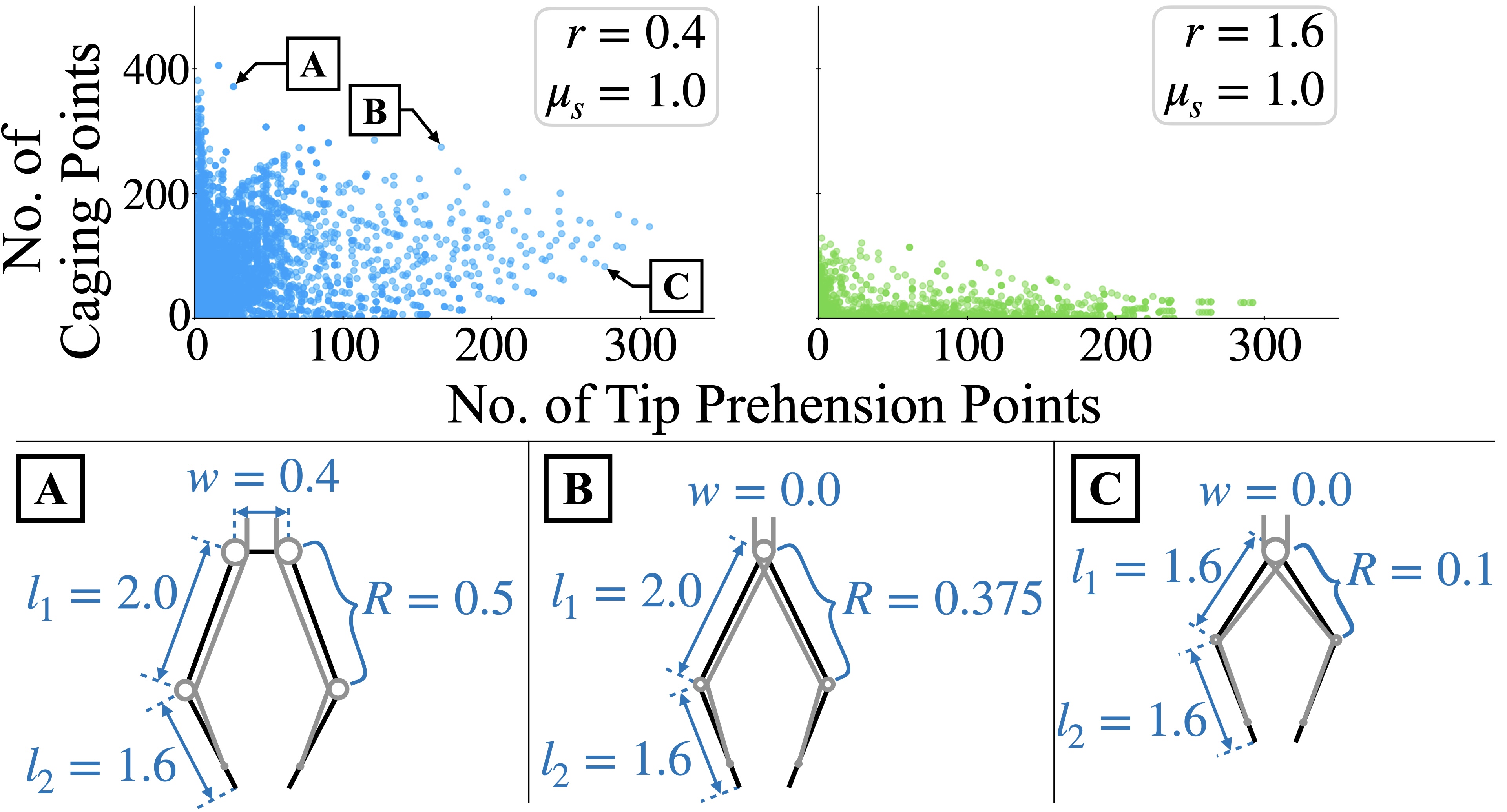}
    \caption[The number of caging points versus the number of tip prehension points for each gripper across two parts. 
    Representative grippers that prioritize caging, tip prehension, or a compromise between the two are shown.]
    {The number of caging points versus the number of tip prehension points is shown on the top for the smallest (blue, $r = 0.4$) and largest (green, $r = 1.6$) parts with the highest coefficient of friction ($\mu_s = 1.0$) we tested. 
    Each point represents a single gripper design. 
    As the part gets larger (green plot), the number of caging points drastically decreases and the number of tip prehension points increases.
    Representative designs that prioritize caging (A), tip prehension (C), and a compromise between the two (B) is shown on the right.}
    \label{figure: caging vs. tip prehension}
\end{figure}

\subsection{Caging Versus Tip Prehension}

An investigation into how the number of caging and tip prehension grasps changed as a function of $\mu_{s}$ and $r$ was performed (see Fig.~\ref{figure: caging vs. tip prehension}). 
As expected, the number of caging grasps decreased (and the number of tip prehension grasps increased) as the part size increased. 
A larger part suggests fewer caging grasp configurations, as the fingers may not be long enough to favorably position themselves around the part. 
Additionally, as $\mu_{s}$ increased, the number of tip prehension grasps increased and the number of caging grasps seemed to decrease. 
Friction prevents a part from being pulled in completely, even when the contact normals point toward the palm, since~\eqref{eq: contact force hull} will be satisfied for more points in $\mathcal{P}_j$ as $\mu_{s}$ increases. 
Additionally, sufficient friction permits more tip prehension grasps even when the contact normals are directed away from the palm. 
Representative designs that focus on caging, tip prehension, and a compromise between the two for a part with $r=0.4$ and $\mu_s = 1.0$ can be seen in Fig.~\ref{figure: caging vs. tip prehension}. 
The designs with more of a focus on caging had a larger transmission ratio ($R$), longer finger lengths ($l_1$ and $l_2$), and a wider palm width ($w$) compared to those with a focus on tip prehension grasps. 
This is consistent with the trends from Fig.~\ref{figure: caging vs. tip prehension}.

\subsection{In-Hand Manipulation Comparison}
\begin{figure}[t!]
    \centering
    \includegraphics[width=0.50\textwidth]{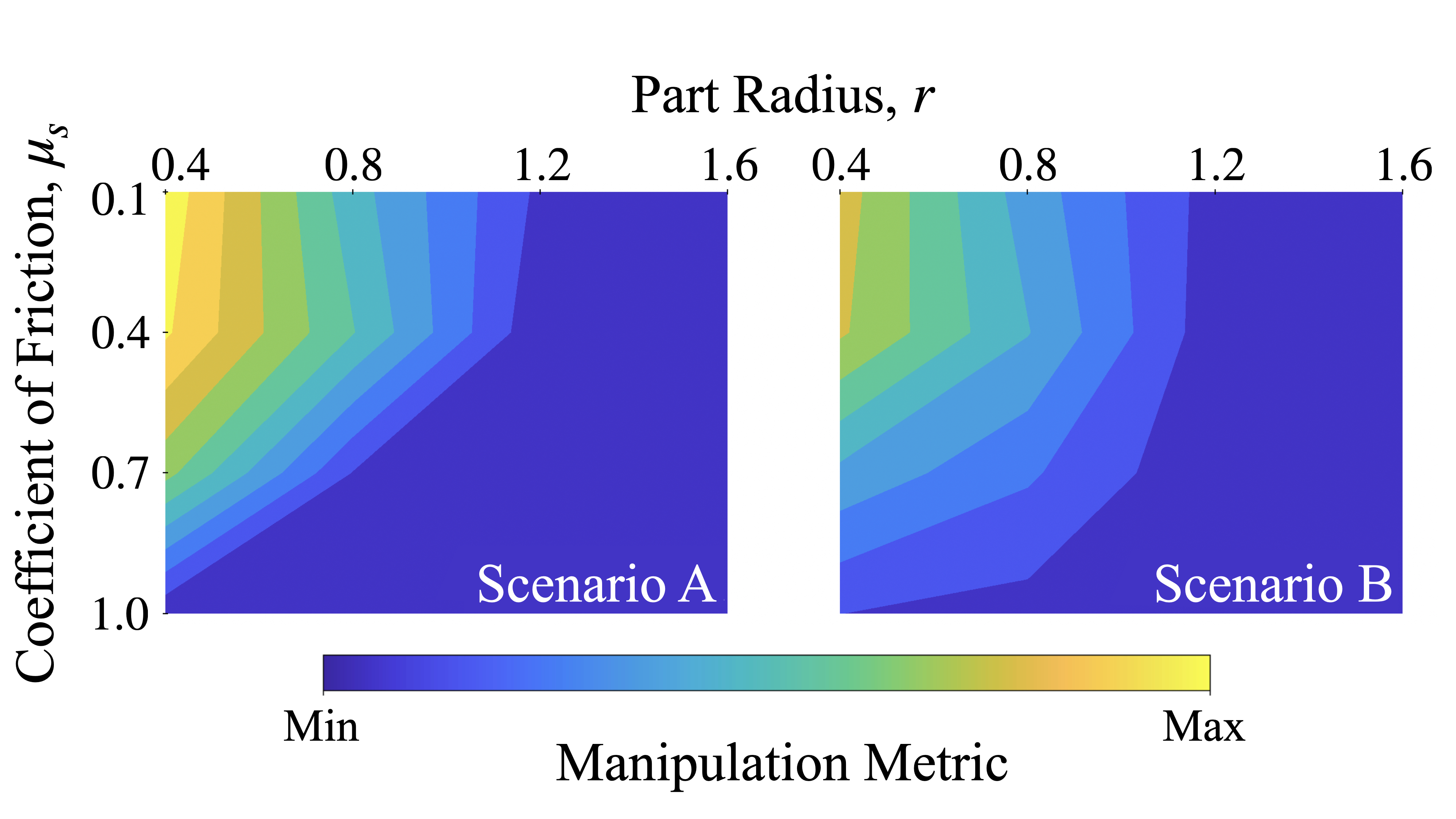}
    \caption{In-hand manipulation metric variation between the best gripper from \textit{Scenario A} and the fully variable gripper from \textit{Scenario B} for all sixteen tested parts.}
    \label{figure: manipulation metric}
\end{figure}

A comparison of \textit{Scenario A} and \textit{Scenario B} is shown in Fig.~\ref{figure: manipulation metric}. 
Across all parts, the best gripper from \textit{Scenario A} had the following parameters: $l_1 = 2.0$, $l_2 = 1.6$, $r_1 = 0.2$, $r_2 = 0.14$.
Recall that $w$ could swing between the full range of values in Table \ref{tab: parameterSearch} for this gripper---this was done because of the results shown in Fig.~\ref{figure: best caging designs}, as we wanted to avoid biasing the results in favor of \textit{Scenario B}.
Parts with $r = 1.6$ and $\mu_s = \{0.7, 1.0\}$ had an in-hand manipulation metric score of zero for both scenarios, most likely a result of the high friction resulting in equilibrium combined with insufficiently long fingers with respect to part size in order to achieve a variety of contact forces.
The two grippers performed more closely than expected. 
Aside from the two parts where both grippers received manipulation scores of zero and despite executing a single actuation command, \textit{Scenario B} performed better for half of the remaining parts compared to \textit{Scenario A}, particularly for parts with higher coefficients of friction.
On the other hand, \textit{Scenario A} performed better for smaller parts and lower coefficients of friction at contact.

\begin{figure}[t!]
    \centering
    \includegraphics[width=0.50\textwidth]{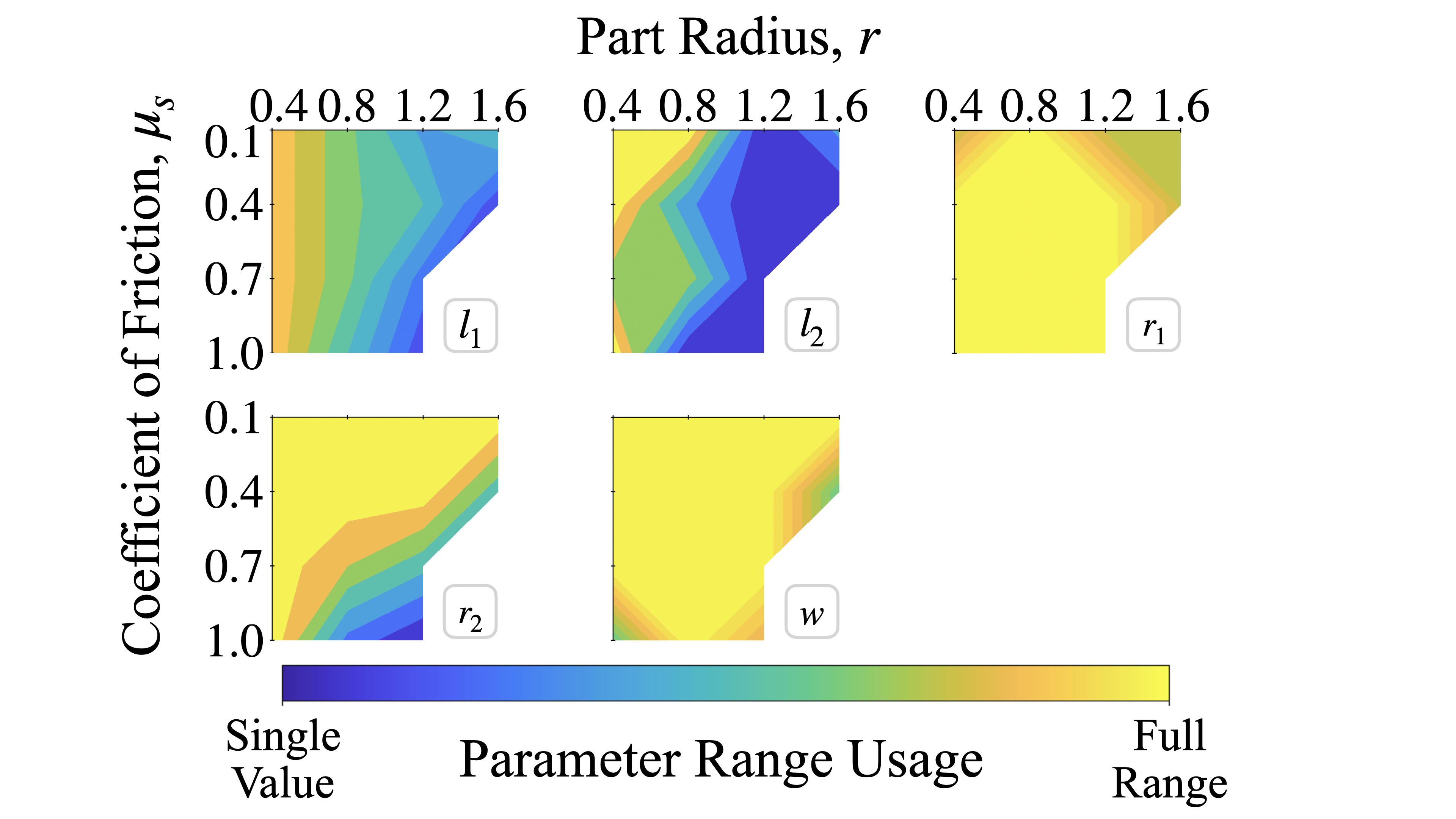}
    \caption[Relative range of a given geometric parameter used in part manipulation.]
    {Relative range of a given geometric parameter used in constructing each convex hull for part manipulation. 
    Note that values resulting in discarded contact forces were not included in this plot.
    Parameter ranges are shown in Table \ref{tab: parameterSearch}.
    Lighter colors indicate parameters that are varied more for in-hand manipulation, while darker colors suggest the parameter has been pushed to a particular value.}
    \label{figure: parameter usage}
\end{figure}

The influence of different parameters for \textit{Scenario B} on the in-hand manipulation metric can be seen in Fig.~\ref{figure: parameter usage}. 
Lighter colors indicate a larger portion of the realizable range for that parameter is needed for in-hand manipulation of a given part. 
Namely, the pulley radii and the palm width utilized their full range of values for almost every part. 
Variable link lengths were less important; as part size increased they were pushed to their upper parameter bounds. 
For smaller parts, the distal link realized its full range, while the proximal link mainly adhered to the upper three-fourths of its parameter range. 
These initial results indicate dynamically changing geometry should primarily be introduced in palm and pulley radii, although link length will play a role regarding greater variation in part size.

\subsection{Takeaways}
Across all experiments, there are several key takeaways.
First, a variable palm width is important for manipulation in general, allowing the fingers to position themselves such that a large variety of resultant contact forces can be achieved, a result also noted by Bircher et al.~\cite{bircher2021complex}.
Additionally, a variable transmission ratio is needed to either draw a part into a cage (large $R$) or maintain a tip prehension grasp (small $R$).
Spanjer et al.~\cite{spanjer2012underactuated} demonstrated this analytically and with a prototype, where increasing $R$ pulled in a part previously held in tip prehension.
A large transmission ratio is also needed to draw in parts when the coefficient of friction at contact is large.
While a variable transmission ratio may be useful, it is mechanically complex and its value must be carefully monitored to prevent phalanx rollback~\cite{birglen2007underactuated}.

\section{CONCLUSION AND FUTURE WORK}
\label{section: conclusion}
This work presented an extension to the energy map method by adding friction to the analysis for two-link underactuated fingers. 
These new energy maps can be used to predict if a grasped part will be caged, ejected, or held in a tip prehension grasp.
Across nearly 6 million grasping scenarios, several design principles were determined.
Namely, a variable palm width is beneficial for caging and manipulating varied part sizes. 
A variable transmission ratio, such as that from Spanjer et al.~\cite{spanjer2012underactuated}, can be used to draw parts into a caging grasp, even in the presence of large friction at contact, as well as to maintain a tip prehension grasp.
While this design is capable, it is also mechanically complex.
Friction modulation may be a suitable alternative, and has previously been used for part translation and rotation~\cite{spiers2018friction, lu2020origami}. 
Changing the friction at contact can modulate the pose of the distal finger relative to the part~\eqref{eq: distal equilibrium position}.
Additionally, it can make a part more manipulable (darker $\rightarrow$ lighter region) or easier to hold in a stable grasp (lighter $\rightarrow$ darker region) as shown in Fig~\ref{figure: manipulation metric}.
In future work, friction modulation at the contact point will be investigated.
Additionally, some experiments should be performed to confirm the practicality of the presented analysis. 

\bibliographystyle{ieeetr}
\bibliography{root}

\end{document}